\documentclass{llncs}

\usepackage{graphicx}
\usepackage{amsmath}
\usepackage{enumitem}
\usepackage{amsfonts}
\usepackage[linesnumbered]{algorithm2e}
\usepackage{subcaption}
\usepackage{tabularx}
\usepackage{xcolor}
\usepackage{capt-of}
\usepackage{multirow}

\title{Extending Dynamic Bayesian Networks for Anomaly Detection in Complex Logs} 
\author{Stephen Pauwels\inst{1} \and Toon Calders\inst{1}}
\institute{University of Antwerp\\ Antwerp, Belgium}

\begin{document}

\maketitle

\begin{abstract}
Checking various log files from different processes can be a tedious task as these logs contain lots of events, each with a (possibly large) number of attributes. We developed a way to automatically model log files with a dozen attributes and detect outlier traces in the data. For that we extend Dynamic Bayesian Networks  to model the normal behavior found in log files. We introduce a new algorithm that is able to learn a model of a log file starting from the data itself. The model is capable of scoring traces even when new values or new combinations of values appear in the log file and has the ability to give a decomposition of the score indicating the root cause for the anomalies.
\end{abstract}

\section{Introduction}
We propose a way of detecting anomalous behavior in Business Processes (BPs). A BP is a series of structured activities in order to perform a task \cite{von2014complete}. Such a sequence of events that together form an instantiation of a BP is called a trace of the business process. In order to monitor a BP, activities are logged in a log file. This file consists of different events and every line in the log file represents a single event. Often log files already indicate which events belong together in the same trace. If not we can apply a clustering algorithm as described in \cite{pauwelsmining} for identifying the different traces.

\begin{example}
\label{ex:logfile}
\emph{The log file in Table \ref{tab:logfile} is generated by a Business Process where an employee needs to log into a system to create a request. This request is then sent to his or her manager who can approve or reject the request. The log consists of 7 attributes: Time, EventID, Type, Activity, UserID, UserName and UserRole. We also keep track of the trace to which an event belongs. In total we have 4 users, each with a unique ID and Name. Every user has a role from a limited set of roles. For the sake of simplicity we have only captured a subset of all possible actions that can occur.}

\begin{table}[t]
\centering
\scriptsize
\begin{tabular} {l l l l l l l | l}
Time & ID & Type & Activity & UserID & UserName & UserRole & tID \\ \hline
0 & 0 & User-Actions & Log in & 001 & User1 & employee & 1\\
1 & 1 & User-Actions & Logged in & 001 & User1 & employee & 1\\
1 & 2 & Request Permission & Create Request & 001 & User1 & employee & 1\\
2 & 3 & Request Permission & Send Mail & 001 & User1 & employee & 1\\
3 & 4 & User-Actions & Log in & 001 & User1 & employee & 2\\
4 & 5 & User-Actions & Logged in & 001 & User1 & employee & 2\\
6 & 6 & Request Permission & Create Request & 001 & User1 & employee & 2\\
7 & 7 & Request Permission & Send Mail & 001 & User1 & employee & 2\\
8 & 8 & Request Permission & Disapproved & 002 & User2 & manager & 2\\
9 & 9 & User-Actions & Log in & 003 & User3 & employee & 3\\
10 & 10 & User-Actions & Logged in & 003 & User3 & employee & 3\\
10 & 11 & Request Permission & Create Request & 003 & User3 & employee & 3\\
11 & 12 & Request Permission & Approved & 002 & User2 & manager & 1\\
12 & 13 & Request Permission & Send Mail & 003 & User3 & employee & 3\\
17 & 14 & Request Permission & Approved & 004 & User4 & sales-manager & 3\\
\textcolor{red}{18} & \textcolor{red}{12} & \textcolor{red}{User-Actions} & \textcolor{red}{Log in} & \textcolor{red}{001} & \textcolor{red}{User1} & \textcolor{red}{manager} & \textcolor{red}{4}\\
\textcolor{red}{19} & \textcolor{red}{13} & \textcolor{red}{User-Actions} & \textcolor{red}{Logged in} & \textcolor{red}{001} & \textcolor{red}{User1} & \textcolor{red}{manager} & \textcolor{red}{4}\\
\textcolor{red}{20} & \textcolor{red}{14} & \textcolor{red}{Request Permission} & \textcolor{red}{Create Request} & \textcolor{red}{001} & \textcolor{red}{User1} & \textcolor{red}{manager} & \textcolor{red}{4}\\
\textcolor{red}{21} & \textcolor{red}{15} & \textcolor{red}{Request Permission} & \textcolor{red}{Approved} & \textcolor{red}{001} & \textcolor{red}{User1} & \textcolor{red}{manager} & \textcolor{red}{4}\\
\textcolor{red}{21} & \textcolor{red}{16} & \textcolor{red}{Request Permission} & \textcolor{red}{Send Mail} & \textcolor{red}{001} & \textcolor{red}{User1} & \textcolor{red}{manager} & \textcolor{red}{4}\\
\end{tabular}
\caption{Example Log file containing normal (black) and anomalous (red) traces}
\label{tab:logfile}
\end{table}
\end{example}

In the context of Business Processes, the detection of anomalous behavior is an important problem. Therefor, in this paper we describe an anomaly detection system that can find deviating traces. This is done by learning the structure and parameters of a model that reflects the normal behavior of a system. Our model takes all attributes and relations between attributes into account, in contrast to existing techniques  \cite{van2005process}. Which provides us with a lot more useful information since log files created by an autonomous system often consist of many more attributes. Attributes can influence each other within an event and between different events. Besides missing activities or a wrong ordering of activities there can be constraints on the activities enforcing that two activities must be performed by the same person or that a person needs to have a certain role to perform an action. 

Diagrams like BPMN models are a great tool for human understanding of a Business Process. For applications such as anomaly detection, BPMN models are, however, insufficiently powerful as they lack the ability to easily express joint probability distributions and multiple attributes; they focus on a single perspective (i.e. the resource-activity perspective). 
Therefore, in order to take advantage of all possible relations between attributes in a log file we create a model based on Dynamic Bayesian Networks (DBNs) \cite{russell2009artificial}. DBNs are an extension of Bayesian Networks that are able to incorporate discrete time. This model will link current events to their predecessors in order to find relations between these events rather than only relations within one event.

In this paper we identify and improve two large shortcomings of DBNs when it comes to modeling the allowable sequences in a log:
\begin{itemize}
\item DBNs are not able to handle unseen values in an appropriate way for business process logs.
\item The case where a value always occurs together with another value describes a common structure in log files. We can model these relations in a DBN but only implicit which may lead to less effective structures.
 \end{itemize}
Therefor we extended the formalism of Dynamic Bayesian Networks to incorporate the aspects that are typical for log files. We will show that our extended Dynamic Bayesian Networks perform well for detecting anomalies. 
 
The structure of our paper is as follows. Section \ref{sec:related_work} describes existing approaches to this (or similar) problems. Section \ref{sec:model} introduces the model for describing normal behavior in log files. We then use this model in Section \ref{sec:anomaly} in order to discover anomalies in traces of events found in log files. The construction of the model is described in Section \ref{sec:learning}. We will evaluate our new method in Section \ref{sec:experiments}.

\section{Related Work}
\label{sec:related_work}
The problem we are interested in is that of finding anomalous sequences (traces) within a large database of discrete multivariate sequences. Different techniques have been proposed to solve this problem both in the anomaly detection field \cite{chandola2012anomaly,nolle2016unsupervised,Bertens2017information,ye2000markov}, as in the process mining field \cite{bezerra2009anomaly,bohmer2016multi}. Some of these techniques use signatures of known anomalies that can occur in the system. It is clear that these systems cannot recognize a new type of anomaly and are too limited for our purpose. We are interested in techniques that build a model, such as Markov Chains that represent normal behavior of a system.

A first type of algorithms works on a database of univariate sequences; i.e., they only take the activity into account. Bezerra et al. \cite{bezerra2009anomaly} investigated the detection of anomalies in a log file using existing Process Mining algorithms in order to build a model of the process. This model is then used to detect anomalous executions of this process. They only use information about the activities performed so they can use standard Process Mining techniques which do not take the extra attributes into account.
Nolle et al. \cite{nolle2016unsupervised} propose an unsupervised anomaly detection method based on neural networks in noisy business process event logs. Using these neural networks makes it possible to reduce the impact of noise in the dataset, where other methods need a training dataset without anomalies as a reference. They do this by adding extra noise to the data before feeding the data into the neural network. Next the neural network is trained to reproduce its input. After the training phase, the network can be used to reproduce the traces from the same input log without the noise. Normal traces are expected to be reproduced with less errors than anomalous traces. 

Other algorithms work on databases of multivariate sequences. Bertens \cite{Bertens2017information} uses MDL to identify multivariate patterns that helps him detect and describe anomalies. A code table consisting of mappings between encodings and frequently occurring patterns is first generated by their algorithm called DITTO \cite{bertens2015keeping}. The anomaly score is defined by dividing the length of the encoded sequence given the code table on the whole dataset by the length of the sequence.
Bohmer et al. \cite{bohmer2016multi} introduce a probabilistic model that is able to score events in the data. First a Basic Likelihood Graph is constructed where all activities are nodes and the edges between nodes indicate the probability that given the previous activity, a certain activity happens next. In the next phase this graph is extended by adding a \emph{resource} and \emph{weekday} between two activities that correspond to the resource that performed the previous action on a particular weekday. Using this graph it is possible to compute a baseline-score given the occurrence of a particular activity. This baseline-score is compared with the actual score given to an execution trace by the model. To score an actual trace Bohmer et al. use the data in the graph with the corresponding probabilities to get a score for the entire trace. Besides data present in the graph, the model is also able to deal with new values. However, they do not describe and test the use of more attributes in detail, but their model can be extended in a straightforward way to other attributes as well. A summary of the different techniques can be found in Table \ref{tab:rel_work}.

\begin{table}[t]
\centering
\begin{tabular}{ | l | c | c | c |}
\hline
& Univariate & Multivariate & Method\\
\hline
Our method  & & \checkmark & Dynamic Bayesian Network\\
Ye \cite{ye2000markov} & \checkmark & & Markov Chains\\
Bezerra \cite{bezerra2009anomaly} & \checkmark & & Process Mining\\
Nolle \cite{nolle2016unsupervised} & \checkmark & & Neural Networks\\
Bertens \cite{Bertens2017information} & & \checkmark & Minimum Description Length\\
Bohmer \cite{bohmer2016multi} & & \checkmark & Probabilistic Model\\
\hline
\end{tabular}
\caption{Summary of Related Work in comparison with our proposed method}
\label{tab:rel_work}
\end{table}

On the topic of handling unseen values Milch et al. \cite{milch2004blog} proposed BLOG, a language to model the generation of a possible world of values. However a well formed language, it still lack some functionality we would like to add to the Dynamic Bayesian Networks.

\section{Extended Dynamic Bayesian Networks}
\label{sec:model}
In this section we extend Dynamic Bayesian Networks to create a model which is more flexible and powerful when dealing with log files. Therefor we first formally define a log file.

\begin{definition}
We assume that $\mathcal{A} = \{A_1, \ldots, A_n\}$, an ordered set of attributes, is given. For each attribute $A_i$ a set of allowed values $dom(A_i)$ is also given. 

An event is a pair $(\mathit{ID}, desc)$ with $\mathit{ID}$ a identifier and desc an event description.  An event description is a tuple $(a_1, \ldots, a_n)$ with $a_i \in dom(A_i)$; $desc.A_i$ denotes $a_i$. We use $e.A_i$ as a shorthand notation for $e.desc.A_i$.

A trace $T = \langle e^1, \ldots, e^i\rangle$ is a sequence of events. A log L is a set of traces, where events in the traces have different identifiers.
\end{definition}

\subsection{History and Context of an event}
To be able to incorporate the timing aspect we introduce the \emph{k-history} and \emph{k-context} of an event.

\begin{definition}
The $k$-history of an event $e^i$ is defined as $\mathcal{H}_k(e^i) = x^k \cdot \ldots \cdot x^1$
where $\cdot$ denotes concatenation and where
\begin{displaymath}
x^l = 
\begin{cases}
e^{i - l}.desc &\text{if $i - l > 0$} \\
(None, \ldots, None) & otherwise\\
\end{cases}
\end{displaymath}
None is a special dedicated value that should not occur in the log.
We use $\mathcal{H}_k(e^i).A^l$ to denote the value of attribute $A$ from the $l$-th event before $e^i$ in the trace that is, $x^l.A$.
\end{definition}

\begin{definition}
The $k$-context of an event $e$ is defined as $\mathcal{C}_k(e) = (\mathcal{H}_k(e) \cdot e.desc)$
we use the notations:
\begin{align*}
\mathcal{C}_k(e).A &:= e.A \\
\mathcal{C}_k(e).A^l &:= \mathcal{H}_k(e).A^l
\end{align*}
\end{definition}

\begin{example}
\emph{For the log in Table \ref{tab:logfile}, the \emph{2-history} of the event with ID 3 is the tuple ($User-Actions^1$, $Logged in^1$, $001^1$, $User1^1$, $employee^1$, $Request Permission$, $Create Request$, $001$, $User1$, $employee$). The \emph{2-context} of this event is the tuple ($User-Actions^2$, $Logged in^2$, $001^2$, $User1^2$, $employee^2$, $Request Permission^1$, $Create Request^1$, $001^1$, $User1^1$, $employee^1$, $Request Permission$, $Send Mail$, $001$, $User1$, $employee$).}
\end{example}

\subsection{Conditional Probability Tables and Functional Dependencies}
In Dynamic Bayesian Networks, the relations within the model are represented using Conditional Probability Tables (CPTs).
\begin{definition}
A $CPT(X | Y)$ is a table where each row contains the conditional probability for a value of $X$ given a combination of values of $Y$.
\end{definition}
The following example indicates the problems we have when using only CPTs for describing BP log files:

\begin{example}
\label{ex:fds}
\emph{Consider the situation where every User has a particular Role and certain activities can only be executed by certain roles. The attribute Role depends on the User and the Activity in this example. When building a single CPT we have to add a row for every possible combination of values for User and Activity, resulting in a large table with a lot of probabilities equal to 1. Also, when a new user is added to the system, all combinations with this user would have to be added to the CPT.} 
\end{example}
To avoid these problems we introduce a new type of relation: a Functional Dependency.

\begin{definition}
Given a log $L$. A Functional Dependency $A^{t_1} \rightarrow B^{t_2}$ holds in $L$ if for all events $e, f \in L$ holds that if $\mathcal{C}(e).A^{t_1} = \mathcal{C}(f).A^{t_1} \not = None$, then $\mathcal{C}(e).B^{t_2} = \mathcal{C}(f).B^{t_2}$ for attributes $A$ and $B$ and time steps $t_1$ and $t_2$.
\end{definition}
A Functional Dependency (FD) between attributes X and Y can be represented by a function $FD_{X\rightarrow Y}: a\_dom(X) \rightarrow a\_dom(Y)$, $FD_{X\rightarrow Y}(x) = y$, with x and y the respective values for attributes X and Y. $a\_dom(A)$ is defined as follows:

\begin{definition}
Let L be a log over $\mathcal{A}$ and $\{A_{i_1}, \ldots, A_{i_k}\} \subseteq \mathcal{A}(L)$. We define the active domain $a\_dom(A_{i_1}, \ldots, A_{i_k}) = \{ (e.a_{i_1}, \ldots, e.a_{i_k}) | \forall T \in L : e \in T \}$ as the set containing all values that occur in the log for the given attributes.
\end{definition}

\begin{example}
\emph{In the log in Table \ref{tab:logfile}, $UserID \rightarrow UserRole$ is a Functional Dependency. Every value of UserID uniquely maps to a value of UserRole. A particular value in UserRole can however occur together with multiple values in UserID. We have the following mappings in our log:}
\begin{displaymath}
\{ 001 \mapsto employee,~002 \mapsto manager,~003 \mapsto employee,~004 \mapsto sales-manager\}
\end{displaymath}
\end{example}

It is possible to mimic this behavior by only using CPTs with all probabilities set to 1. Introducing Functional Dependencies, however, allows us to create easier models that can be used to express more general patterns. Furthermore we are able to first check for these FDs, making the search for the Conditional Dependencies less complex. When using separate FDs we are also able to determine the exact FD for which we got deviating values, CPTs do not provide us with the same amount of expressiveness.

A second major shortcoming of CPTs when dealing with log files is that only values that have occurred in the training dataset will be present in the tables. In a log file it might be normal for new users to appear without these events being anomalous. The model will assign a probability of $0$ to these values. If the new value, however, satisfies all other relations then it is likely to be a correct event. Smoothing could be used, but may be inappropriate for attributes with frequent new values. The frequency of new values depends on the attribute itself, not on the log file.


\subsection{Extending the Dynamic Bayesian Networks}

Combining all these elements, we extend the definition of a DBN as follows:

\begin{definition}
An extended DBN with memory $k$ over $\mathcal{A}$ is a tuple:
\begin{equation*}
(G, FDR, \mathcal{CPT}, \mathcal{FD}, new\_value, new\_relation, violation):
\end{equation*}
\begin{itemize}
\item $G(V,E)$ is a directed acyclic graph with $V = \mathcal{A}^k \cup \ldots \cup \mathcal{A}^1 \cup \mathcal{A}$ where $\mathcal{A}^i = \{A^i | A \in \mathcal{A} \}$ for $i = 1, \ldots, k$, and $E \subseteq V \times \mathcal{A}$.\\
$\mathcal{A}^i$ represents the attributes of the $i$th event before the current event.\\
$E$ expresses dependencies of the attributes of the current event on the other attributes in its context.
\item $FD \subseteq E$ denotes the set of dependencies that are functional.
\item For each variable $A \in \mathcal{A}$, $Parents(A)$ denotes the set of variables\\ $\{ B \in V | (B,A) \in E \setminus FD\}$.
\item $\mathcal{CPT}$ consists of a Conditional Probability Table $CPT(A | Parents(A))$ for each $A \in \mathcal{A}$
\item $\mathcal{FD}$ consists of a Mapping $FD_{A \rightarrow B}$ for each $(A,B) \in FD$
\item $new\_value(A) \text{ is a function } \mathcal{A} \rightarrow [0,1] \text{ representing the probability of encountering unseen values }$
\item $new\_relation(A) \text{ is a function } \mathcal{A} \rightarrow [0,1] \text{ representing the probability of encountering an unseen}\\ \text{combination of parent values for the CPT.}$
\item $violation(X,Y) \text{ is a function } \mathcal{A} \times \mathcal{A} \rightarrow [0,1] \text{ representing the probability that $FD_{X \rightarrow Y}$ is broken.}$
\end{itemize}
\end{definition}

\begin{figure}[h]
\centering
\includegraphics[scale=0.30]{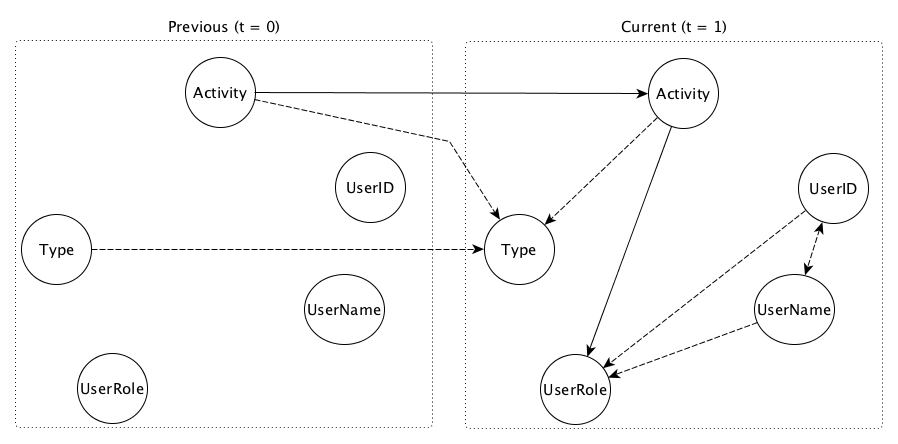}
\caption{eDBN with conditional (full) and functional dependencies (dotted)}
\label{fig:edbn_example}
\end{figure}

Figure \ref{fig:edbn_example} shows a possible eDBN based on our example.

\label{ex:model}

\subsubsection{The joint distribution of an eDBN}
An eDBN with memory $k$ represents a joint distribution over sequences as follows:
\begin{align}
P(\langle e^{1}, \ldots, e^m \rangle) &= \prod_{e \in \langle e^{1}, \ldots, e^m \rangle}{P(e | \mathcal{H}_k(e))}\\
&= \prod_{e \in \langle e^{1}, \ldots, e^m \rangle}{\prod_{A \in \mathcal{A}}{P(e.A | \left.\mathcal{C}_k(e)\right|_{Parents(A)})}}
\end{align}

The probability for an attribute in an event consists of three different parts.
The first part checks for new values and is defined as:
\begin{equation}
\scriptsize
value_A(x)=
\begin{cases} 1 - new\_value(A) & \text{if } x \in a\_dom(A)\\
new\_value(A) & \text{otherwise}
\end{cases}
\end{equation}

The probability for the Conditional Dependency is given as follows:
\begin{equation}
\scriptsize
Relation(x_i | Parents(X_i)) =
\begin{cases}
new\_relation(Parents(X_i) & \text{if new combination of parent values.}\\
(1 - new\_relation(Parents(X_i)) * \\~~~~~~~~~~~~~~~~CPT(x_i | Parents(X_i)) &\text{otherwise}
\end{cases}
\end{equation}

The probability for a Functional Dependency is expressed as follows: 
\begin{equation}
\scriptsize
FDM_{X,Y}(y | x) = 
\begin{cases}
1 - violation(X,Y) & \text{if } FD_{X\rightarrow Y}(x) = y \text{ or } x \not \in a\_dom(X)\\
violation(X,Y) & \text{otherwise}
\end{cases}
\end{equation}

To incorporate all the new elements we introduced in our model we extend the way of determining the probability in contrast to original BNs.
\begin{align*}
\scriptsize
P(e.A | \left.\mathcal{C}_k(e)\right|_{Parents(A)}) = value_A(e.A) \cdot &Relation(e.A | Parents(A))\\
						&\cdot \prod_{(X, A) \in FDR}{FDM_{X\rightarrow A}(e.A | \mathcal{C}_k(e).X )})
\end{align*}

\begin{example}
\emph{The probability for an event $e$ in the model given in Figure \ref{fig:edbn_example} is equal to:}
\begin{align*}
\scriptsize
& value(Activity^1) Relation(Activity^1 | Activity^0) value(Type^1) \\
& \cdot FDM(Type^1 | Activity^0)FDM(Type^1 | Activity^1) FDM(Type^1 | Type^0)\\
& \cdot value(UserID^1) FDM(UserID^1 | UserName^1)value(UserName^1)\\
& \cdot FDM(UserName^1 | UserID^1) value(UserRole^1)Relation(UserRole^1 | Activity^1)\\
& \cdot FDM(UserRole^1 | UserID^1) FDM(UserRole^1 | UserName^1)\\
\end{align*}
\emph{The value for the attribute $UserRole_1$ for the event with ID 2 is:}
\begin{align*}
		&value(UserRole^1) Relation(UserRole^1 | Activity^1)\\
		& ~~~~~~* FDM(UserRole^1 | UserID^1) FDM(UserRole^1 | UserName^1)\\
		& = (1 - 0.2) * (1 - 0.4) * 1 * (1 - 0) * (1 - 0) = 0.48\\
\end{align*}
\end{example}

This score can be decomposed to find the root cause for any anomaly in the data. This will be further elaborated in future work.

\subsection{Anomaly detection}
\label{sec:anomaly}
To find anomalous sequences of events we use a score-based approach. The score is obtained by calculating the probability for a trace $\langle e^1, \ldots, e^n\rangle$ given a model $m$. We normalize the result using the $n$-th root, with $n$  the number of events in the trace. This normalization makes sure that longer traces are not penalized.
\begin{equation}
Score( \langle e^1,\ldots,e^n\rangle) = \sqrt[\leftroot{-1}\uproot{6}n]{P(\langle e^1, \ldots, e^n \rangle)}
\end{equation}
Sequences with a high score thus have a high probability of occurring and are most likely to represent normal behavior, whereas low scores indicate higher chances of being an anomaly. We return a sorted list of traces, sorted by their scores. The idea is that a user can only handle the first $k$ anomalies detected. Since we can score any sequence of events, we do not have to wait for a complete trace before we can score it. The model can thus be used to detect anomalies in ongoing traces.

\section{Learning the structure and parameters of the model}
\label{sec:learning}
We build our model using a reference dataset containing only the normal execution of the process. Our experiments show that the performance of our algorithm is, however, not influenced when the dataset contains a small amount of noise. In order to incorporate the timing aspect we replace every event in the log with its \emph{k-context}. We refer to this log as the \emph{k-context log}.

We can use the k-context log as input for traditional Bayesian Network learning algorithms that have no specific knowledge about the different time steps to find the conditional probability tables. Afterwards we interpret the different attributes in their appropriate time slice. The complete algorithm for computing the structure can be found in Algorithm \ref{algo:mining}.

\SetKwProg{Fn}{Function}{}{}
\begin{algorithm}
\caption{Algorithm for learning the structure and parameters of eDBNs}
\Fn{LearnEDBN} {
	\KwData{variables, FDThreshold}
	\KwResult{The learned eDBN}
	V = $vars$\\
	FD = $\{ X \rightarrow Y : X,Y \in V~|~U(X|Y) > FDThreshold\}$ \\
	blacklist = $\{ X \rightarrow Y: \forall X \in V_i, Y \in V_j~with~i \geq j > 0\}$\\
	whitelist = FD\\
	G(V, E) = LearnBayesianNetwork(variables = V, blacklist, whitelist) \\
	FDS = ConstructFunctionalDependencyFunctions(FD)\\
	CPT = ConstructConditionalProbabilitiesTables($E \setminus FD$)\\
	NV = $\{ X \mapsto \frac{|a\_dom(X)|}{|~L~|}: \forall X \in V  \}$\\
	NR = $\{ X \mapsto \frac{|a\_dom(Parents(X))|}{|~L~|}: \forall X \in V \}$\\
	VIOL = $\{ X \times Y \mapsto \frac{ | \{  e \in L : FD_{X\rightarrow Y}(e.X) \neq e.Y \} | }{|L|}: \forall (X,Y) \in FD \}$\\
	\Return{eDBN(G(V, $E \setminus FD$), FD, CPT, FDS, NV, NR, VIOL)}
}

\label{algo:mining}
\end{algorithm}
First the algorithm searches for Functional Dependencies in the data. In order to discover them, the Uncertainty Coefficient \cite{press2007numerical} is applied to the k-context log, which is defined as follows for the random variables $X$ and $Y$:
\begin{align*}
U(X|Y) = \frac{I(X;Y)}{H(X)} \enspace,
\end{align*}
with $H(X)$ the \emph{entropy} \cite{Shannon2001} of $X$ and I(X;Y) the \emph{Mutual Information} \cite{cover2012elements} given as:
\begin{align*}
I(X;Y) =& \sum_{y\in a\_dom(Y)} \sum_{x\in a\_dom(X)} p(x,y)\log{\frac{p(x,y)}{p(x)p(y)}}\\
H(X) =& - \sum_{x\in a\_dom(X)} p(x)log(p(x))
\end{align*}
The Uncertainty Coefficient is the normalized form of Mutual Information. It gives information about how much the values of an attribute depend on another attribute. We use it to determine what attributes are related to each other and how much they relate to each other. The measure ranges from 0 (no correlation between the two attributes) to 1 (completely correlated attributes, thus indicating the existence of a Functional Dependency) \cite{white2004performance}. If $U(X | Y) > threshold$, we will assume that the FD $Y \rightarrow X$ holds. This threshold has to be chosen according to the amount of noise in the data. A higher threshold means a more strict Functional Dependency is used that is less able to cope with noise.

\begin{example}
\emph{When we calculate the Uncertainty Coefficient of $\mathit{UserID_1}$ and $\mathit{UserID_0}$ we get: $U(\mathit{UserID_1} | \mathit{UserID_0}) = \frac{I(\mathit{UserID_1}; \mathit{UserID_0})}{H(\mathit{UserID_1})} = \frac{0.5597}{1.1369} = 0.4923$}
\end{example}

For an attribute A in log L, $new\_value(A)$, $new\_relation(A)$ and $violation(X,Y)$ are defined as follows:
\begin{align*}
new\_value(&A): \frac{|a\_dom(A)|}{|~L~|},~~
new\_relation(A): \frac{ |a\_dom(Parents(A))| }{|~L~|}\\
&violation(X,Y): \frac{ | \{  e \in L : FD_{X\rightarrow Y}(e.X) \neq e.Y \} | }{|L|}
\end{align*}

\begin{example}
\emph{The New Value Rate of the \emph{UserRole} is equal to $\frac{3}{15} = 0.2$. The rate for \emph{Activity} is equal to $\frac{6}{15} = 0.4$. This indicates that new values are more likely to occur for the attribute \emph{Activity} than for \emph{UserRole} according to our data.}
\end{example}

With a standard Bayesian Network learning algorithm we can discover the Conditional Dependencies present in the data. It is possible to use any learning algorithm that uses data to learn its structure. We choose to use a Greedy algorithm that finds a local optimum for the Akaike Information Criterion (AIC) \cite{akaike1974new}. 

The relations present in our model should only indicate a causality relation; events in the present cannot influence events in the past. Therefore edges that do not represent a causality relation are blacklisted. This blacklist is created by adding all edges that do not end in the \emph{current} time step. 

We do not want the algorithm to find edges already labeled as FDs, therefor we add these edges to a whitelist. The Bayesian Net learning algorithm should always include the edges from the whitelist in the model. This way the learning algorithm takes advantage of the information we already know about these FDs.

After running the greedy algorithm we have found the Conditional and Functional Dependencies that define the structures present in our data. We can then combine them into one single model. This gives us the structure of the eDBN-model. The next step in building the model is filling in all the different Conditional Probability Tables (CPTs) and constructing the Functional Dependency functions for all nodes.

\section{Experiments}
\label{sec:experiments}
To properly test our newly proposed method we use two different datasets. The first dataset is a synthetically generated multi-dimensional dataset. The second is the BPI Challenge 2015 (BPIC15) \cite{bpic15} data. This data consists of applications for building permits in 5 Dutch municipalities, we refer to these as BPIC1 to BPIC5, Table \ref{tab:bpic_schema} summarizes the data. We use this last dataset in two different forms: the dataset with anomalies introduced and the dataset with a reduced subset of attributes with anomalies included. We included the same amount of anomalies using a similar approach as described by Bohmer et al. \cite{bohmer2016multi} to best compare our approach.

\begin{table}[b]
\centering
	\begin{tabular}{| c | c | c | c |}
	\hline
	Dataset & Number of traces & Average trace length & Number of Activities\\
	\hline
	BPIC1 & 1199 & 43.5 & 398\\
	BPIC2 & 832 & 53.3 & 410\\
	BPIC3 & 1409 & 42.3 & 383\\
	BPIC4 & 1053 & 44.9 & 359\\
	BPIC5 & 1156 & 51.0 & 389\\
	\hline
	\end{tabular}
\caption{Description of the BPIC datasets}
\label{tab:bpic_schema}
\end{table}

We use the synthetic dataset to test the overall performance of our algorithm, where we try different ratios of anomalies in both training and test set. Next we perform an in-depth comparison with the Likelihood Graphs proposed by Bohmer et al. \cite{bohmer2016multi} using the reduced subset of the BPIC15 data. Furthermore we compare our approach to a variety of algorithms available in the ELKI - tool \cite{SchubertKEZSZ15}, using both the synthetic data and the reduced BPIC15 data. The ELKI - tool contains most of the existing anomaly detection algorithms in a uniform way. The Area Under the Curve (AUC) is used to compare the algorithms. All code used to perform the experiments and generate the datasets can be found on our GitHub repository\footnote{https://github.com/StephenPauwels/edbn}.

\subsection{Testing with synthetic data}
We built a data generation tool that allows us to create log files containing different relations between events. In order to do so we first create a model of sequential activities with depending attributes. The model is based on a BP for shipping goods. Goods can have a value and an extra insurance can be taken. Goods with an extra insurance need a different workflow from goods without extra insurance. The data consists of 13 attributes. We create one model for normal execution and one model for anomalous execution, where we explicitly changed the order of events or use the wrong flow of events according to the insurance chosen. Next we introduce some extra attributes where some of these attributes depend on other attributes. For the anomalous traces we added random values on random places. We generated multiple set-ups with a variable number of anomalies in both training and test data. We added anomalies in our training data to check and show that our approach does not require a flawless log file as training data but is able to deal with a small amount of unexpected behavior in the data.

\begin{table}[t]
\centering
	\begin{tabular}{| c | c || c | c | c | c | c | c | c | c |}
	\hline
	 && \multicolumn{8}{ c |}{Test set}\\
	\hline
	&\% Anomalies & ~0.1~ & ~0.5~ & ~1.0~ & ~2.5~ & ~5.0~ & ~10.0~ & ~25.0~ & ~50.0~ \\
	\hline
	\hline
	\multirow{4}{*}{Training set}& 0.0 & 1.00 & 0.99 & 0.98 & 1.00 & 0.93 & 0.98 & 0.99 & 0.97 \\
	&0.5 & 1.00 & 1.00 & 0.99 & 0.97 & 0.99 & 0.99 & 0.98 & 0.96\\
	&1.0 & 1.00 & 0.99 & 1.00 & 0.99 & 0.98 & 1.00 & 0.98 & 0.97\\
	&2.5 & 1.00 & 1.00 & 0.98 & 1.00 & 0.99 & 1.00 & 0.89 & 0.96\\
	\hline
	\end{tabular}
\caption{AUC values for different combinations of anomalies.}
\label{tab:shipment_auc}
\end{table}

The AUC-scores for different amounts of anomalies in both training and test data can be found in Table \ref{tab:shipment_auc}. This test shows that our algorithm is able to find the relations mentioned in Section \ref{sec:model}, even when the training set contains a small amount of noise or anomalies. 

\subsection{Comparison}
\subsubsection{Comparison with Likelihood Graphs}
In order to compare our approach to the solution presented by Bohmer et al. we first implemented the algorithm found in \cite{bohmer2016multi}. Next we generated data as described by Bohmer et al. starting from the reduced BPIC data. Therefor we randomly split the original data in two equal data sets, one for training and one for testing. In the test data we introduced anomalies according to the description in Bohmer et al. The statistics for the generated files can be found in Table \ref{tab:stats_bpi}. Normal input will, however, never contain this many anomalies.

\begin{table}[t]
\centering
	\begin{tabular}{ | l | c | c | c |}
	\hline
	File & Training size & Test size & Number of anomalies in Test set\\
	\hline
	BPIC1 & 589 & 610 & 291 (47.7\%)\\
	BPIC2 & 408 & 423 & 214 (50.5\%)\\
	BPIC3 & 723 & 686 & 356 (51.8\%)\\
	BPIC4 & 522 & 530 & 257 (48.4\%)\\
	BPIC5 & 595 & 561 & 283 (50.4\%)\\
	\hline
	\end{tabular}
	\caption{Number of traces present in the different log files.}
	\label{tab:stats_bpi}
\end{table}

\begin{figure}[t]
	\centering
	\begin{subfigure}[b]{0.32\textwidth}
		\includegraphics[width=\textwidth]{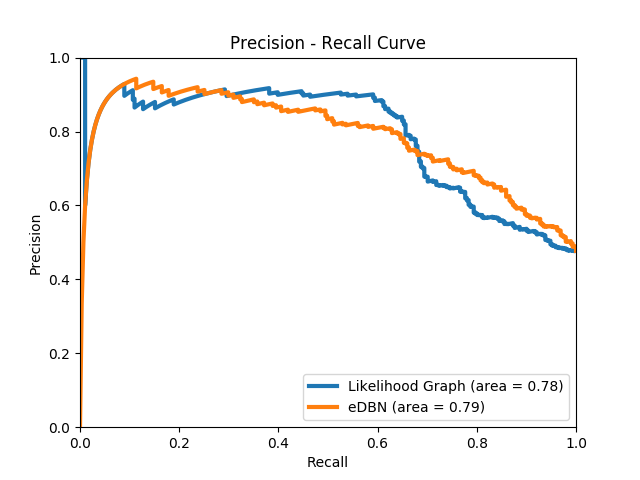}
		\caption{BPIC1}
	\end{subfigure}
	\begin{subfigure}[b]{0.32\textwidth}
		\includegraphics[width=\textwidth]{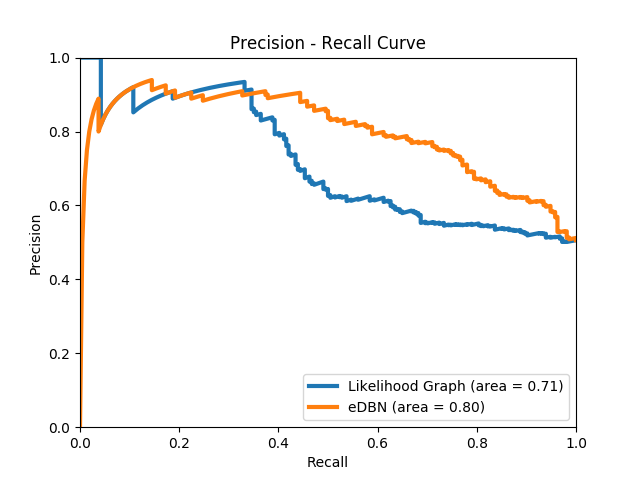}
		\caption{BPIC2}
	\end{subfigure}
	\begin{subfigure}[b]{0.32\textwidth}
		\includegraphics[width=\textwidth]{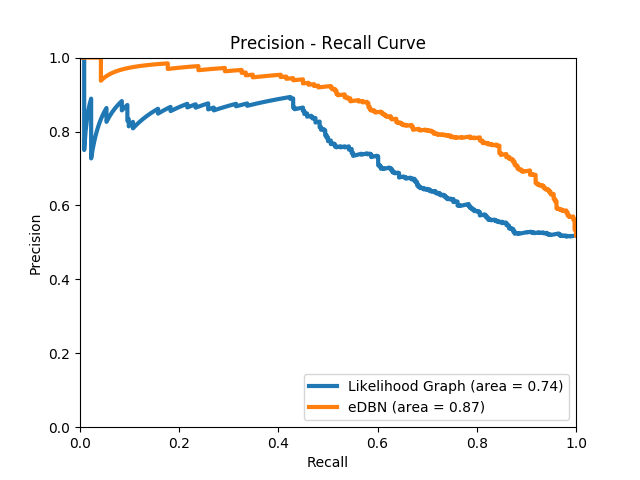}
		\caption{BPIC3}
	\end{subfigure}
	\begin{subfigure}[b]{0.32\textwidth}
		\includegraphics[width=\textwidth]{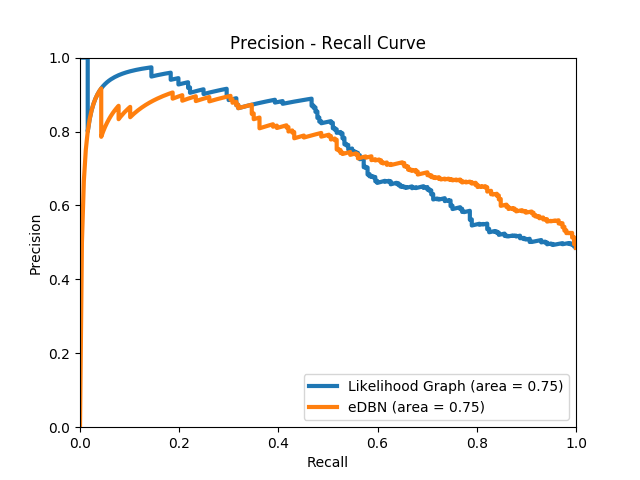}
		\caption{BPIC4}
	\end{subfigure}
	\begin{subfigure}[b]{0.32\textwidth}
		\includegraphics[width=\textwidth]{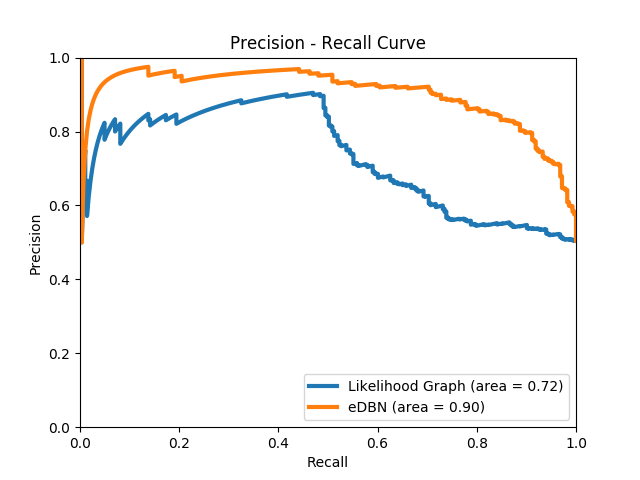}
		\caption{BPIC5}
	\end{subfigure}
	\begin{subfigure}[b]{0.32\textwidth}
		\includegraphics[width=\textwidth]{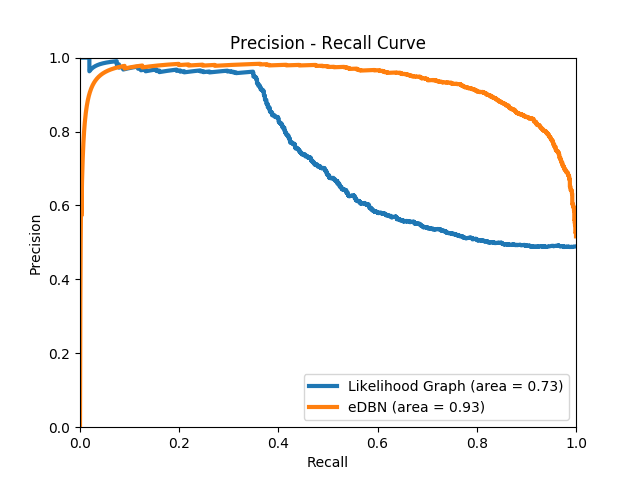}
		\caption{Total BPIC}
	\end{subfigure}
	\caption{Comparison of precision/recall graphs.}
	\label{fig:prec_recall_comparison}
\end{figure}

The Likelihood graph calculates the likelihood for the ongoing trace and compares this with a baseline score in order to indicate if a trace is an anomaly. Since our method works with giving scores and sorting all anomalies, we used the minimum of the difference between the Ongoing Likelihood and the Minimum Likelihood for the different activities within a trace in order to best capture the ideas of the Likelihood Graphs. The lower the difference the more likely it is that this trace contains an anomaly. We used the precision/recall curve to compare the two approaches. The results can be found in the graphs in Figure \ref{fig:prec_recall_comparison} for each of the five different municipalities. Since all five municipalities have different ways of performing the different processes we also created one file containing all data of all municipalities. Then we introduced anomalies in the same way as we did for the other files. This combined dataset allows us to test how well each approach can handle different processes in a single log file.

We can see that the two methods mostly perform equally on the BPIC2015 data. When we look at the results for the combined log file, we can see that we clearly outperform the Likelihood Graph. We can conclude that our model is capable of performing well especially with multiple processes in the log file.

\subsubsection{Comparison with other anomaly detection methods}
We also tested our method against other anomaly detection methods (not necessarily methods that take into account the sequential nature). We used the \emph{k-context} format as input for all the algorithms. The best parameters were chosen after performing some experiments. We performed the experiments using the ELKI - tool \cite{SchubertKEZSZ15}. Since none of these methods uses a different (clean) training dataset we used the same file to generate our model as to test the model. The results can be seen in Table \ref{tab:anoms_table}. We see that only ALOCI outperforms us on the BPIC data, this is due to the fact that we used the same file for training and testing. Our method and Bohmer et al. performs best when having a clear training dataset. For some algorithms we were not able to get results for both datasets (due to runtimes and memory usage).
\begin{table}[t]
\centering
\scriptsize
	\begin{tabular}{| l | c | c | c |}
		\hline
		& \multicolumn{2}{c |}{AUC} & \\
		\hline
		Method & Synth data & BPIC data & Remark\\
		\hline
		eDBN & 1.00 & 0.84 & ~no FDs were found for BPIC~\\
		eDBN without FD & 0.69 & 0.84 & \\
		Bohmer et al. \cite{bohmer2016multi} & 1.00 & 0.58&\\
		FastABOD \cite{kriegel2008angle} & 0.50 & 0.56&\\
		LOF \cite{breunig2000lof} & 0.49 & 0.55&\\
		SOD \cite{kriegel2009outlier} & 0.53 & 0.60&\\
		Feature Bagging \cite{lazarevic2005feature} & 0.75 & 0.57 & \\
		SimpleCOP \cite{zimek2009correlation} & 0.53 & 0.60&\\
		LibSVMOneClassOutlierDetection \cite{scholkopf2001estimating} & 0.51 & 0.46&\\
		COP \cite{kriegel2012outlier} & 0.81 & 0.63&\\
		DWOF \cite{momtaz2013dwof} & 0.51 & 0.44&\\
		OpticSOF \cite{breunig1999optics} & 0.53 &0.46&\\
		ALOCI \cite{papadimitriou2003loci} & - & 0.86&\\
		Bertens et al. \cite{Bertens2017information} & - & - & not able to get a complete run\\
		\hline
	\end{tabular}
\caption{Overview of results for different Anomaly detection techniques. }
\label{tab:anoms_table}
\end{table}

\section{Conclusion}
In this paper we extended Dynamic Bayesian Networks in order to create a new model that allows us to better and in more detail describe the structure and properties of a log file generated by process-aware information systems. As standard DBNs have shortcomings for analyzing these logs we added some elements to cope with these shortcomings. We added Functional Dependencies for a better description of the structure of a log file. Since DBNs cannot cope with unseen values we also improved the way our model deals with these unseen values. Next we described our algorithm for creating models that reflect the multidimensional and sequential nature of log files. We conducted different types of experiments: the first experiment confirmed that our algorithm achieves high performance in different settings with different amounts of anomalies in both training and test sets. Next we compared our approach with existing solutions. 
In the future we would like to extend our model in order to incorporate the time aspect even better by introducing an extra timing element that is capable of dealing with duration of activities and time gaps between activities.

\bibliographystyle{splncs}
\bibliography{Paper}

\end{document}